\documentclass[10pt,twocolumn]{article}

\usepackage[utf8]{inputenc}
\usepackage[letterpaper,margin=0.75in,footskip=0.375in]{geometry}
\usepackage{hyperref}
\usepackage{amsmath}
\usepackage{amssymb}
\usepackage{graphicx}
\usepackage[font=footnotesize,labelfont=bf,labelsep=period]{caption}
\usepackage{booktabs}
\usepackage{subcaption}
\usepackage[natbibapa]{apacite}

\usepackage[table]{xcolor}

\definecolor{LightGray}{gray}{0.95}

\makeatletter % change macro w/ @ in name
\newcommand{\subpara}[1]{
    \vspace{4mm}\noindent\textbf{#1}\hspace{0.5mm}
    \@ifnextchar\par\@gobble\relax
}
\makeatother

\AtBeginDocument{\urlstyle{APACsame}}
\hypersetup{colorlinks=true,urlcolor=blue,citecolor=blue,linkcolor=blue,pdftitle={SIAM}}

\title{SIAM: Head and Brain MRI Segmentation from  \\ Few High-Quality Templates via Synthetic Training}

\author{
    \textbf{Romain~Valabregue}\textsuperscript{a,*},
    \textbf{Ines~Khemir}\textsuperscript{a}, 
    \textbf{Eric~Badinet}\textsuperscript{a}, \\
    \textbf{François~Rousseau}\textsuperscript{b},
    \textbf{Guillaume~Auzias}\textsuperscript{c, \dag}
    \textbf{Reuben~Dorent}\textsuperscript{a,\dag}
}

\date{
    \footnotesize
    $^a$ Sorbonne Université, Institut du Cerveau - Paris Brain Institute \\  ICM, CNRS, Inria, Inserm, AP-HP, Hôpital de la Pitié Salpêtrière, F-75013, Paris, France \\
    $^b$ IMT Atlantique, LaTIM INSERM U1101, Brest, France  \\
    $^c$ Aix-Marseille Université, CNRS, Institut de Neurosciences de la Timone, UMR 7289, Marseille, France  \\
    \dag Equal contribution.\\
    $^*$ Corresponding author, \href{mailto:mhoffmann@mgh.harvard.edu}{romain.valabregue@upmc.fr}. \\
}

\begin{document}

% Abstract.
\twocolumn[
\begin{@twocolumnfalse}
    \maketitle

    \begin{abstract}
  Synthetic training has recently advanced brain MRI segmentation by enabling contrast‑agnostic models trained entirely on generated data. However, most existing approaches rely on hundreds of automatically labeled templates, introducing systematic biases and limiting their flexibility to incorporate new anatomical structures. 
  We present the Segment It All Model (SIAM), a 3D whole-head segmentation framework for 16 anatomical structures, trained using only six high‑quality, manually annotated templates. SIAM extends domain randomization to both intensity and shape domains: synthetic image generation ensures contrast variability, while high‑resolution spatial transformations model anatomical differences in cortical thickness and deep nuclei morphology.
  Unlike prior synthetic models, SIAM simultaneously segments brain as well as extra‑cerebral tissues, including cerebrospinal fluid, vessels, dura mater, skull, and skin, enabling fully automated, preprocessing‑free analysis. Evaluation across eight heterogeneous datasets (N=301), that include multiple contrasts (T1‑weighted, T2‑weighted, CT) and span a wide range of ages, demonstrates that SIAM matches or outperforms state‑of‑the‑art methods for brain structures, in addition to extending automated segmentation to non-brain structures. The model also exhibits superior consistency across contrasts and repeated acquisitions, together with improved sensitivity to subtle gray matter atrophy. We openly release the model and the label templates at \url{https://github.com/romainVala/SIAM}.
  
        \subpara{Keywords} Segmentation, synthetic training, domain randomization, contrast agnostic, brain, skull, head, vessel   \\
        \ \\
        \textbf{Abbreviations} GM: cortical Gray Matter. WM: White Matter. CSF: Cerebrospinal fluid. REF: Reference Annotation. DR: Domain Randomization.
        
    \end{abstract}

    \vspace{2\baselineskip}
\end{@twocolumnfalse}
]

\section{Introduction}

The segmentation of anatomical tissues from brain Magnetic Resonance Imaging (MRI) is a critical task in medical image analysis. It is essential for the design of imaging biomarkers such as for cortical gray matter and deep nuclei atrophy. It is also a key component of most MRI preprocessing pipelines, where it serves to define anatomically relevant regions for downstream analyses. 

Various methods exist for segmenting brain anatomy. Classical methods (FreeSurfer, SPM, FSL, ANTs, SAMSEG) are widely used and rely on fitting a multi-Gaussian distribution within a spatial prior. Each tool has its own biases and the results depend on the quality of the data preprocessing \cite{tustison_large-scale_2014,ashburner_unified_2005}. Recently, deep learning segmentation models have shown promising results, mitigating the need for data preprocessing. They rely on supervised training, either on a small number of subjects with manually defined labels \citep{coupe_assemblynet_2020,huo_3d_2019} or on larger samples but with automated segmentation serving as ``silver-standard'', typically from FastSurfer \citep{henschel_fastsurfervinn_2022,svanera_fighting_2024}, or a combination with fine tuning on manual labels \citep{roy_bayesian_2019,wachinger_deepnat_2018}. 

Defining labels from automated tools limits the segmentation task to structures already available, without the possibility to add new labels or to improve their quality. Another key limitation is the poor generalization toward new contrasts of the proposed models. Most training are performed on T1w images, and intensity data augmentations are too limited to make the models robust to large contrast changes \cite{isensee_automated_2019, valabregue_comprehensive_2024}. 

A notable approach for improving model robustness to new contrasts is the \emph{synthetic training} method, originally proposed by \cite{billot_synthseg_2023}. This approach achieves contrast-agnostic performance, thereby addressing generalization issues due to intensity variability. Its success relies on two key elements: (i) an explicit image-generative model starting from label templates, and (ii) the principle of Domain Randomization (DR). DR states that one does not need to reproduce realistic distributions; instead, introducing random variations larger than those expected in real data improves generalization.

Applying DR in the intensity domain is straightforward. Since synthetic images are rendered from a direct signal model, tissue intensities can be sampled within a normalized rank (e.g. [0,1]). This randomization of the contrast is the key to obtaining contrast-agnostic models and can be viewed as an extreme intensity augmentation as the real data is no longer needed. Another advantage is for evaluation, real test datasets can be considered as out-of-distribution, which strengthens our confidence in the evaluation results.

The SynthSeg model demonstrated the effectiveness of this contrast agnostic property in real-world applications \citep{billot_synthseg_2023}. In this work, we investigate an additional benefit of synthetic training: its capacity to produce unbiased models, resulting in more accurate volumetric measurements. Indeed, another key advantage of synthetic training is its reduced sensitivity to labeling errors. Since synthetic images are generated directly from the templates, there is by design a perfect consistency between templates and images, unlike manual or automatic annotations, which inevitably contain labeling inaccuracies. In principle, this relaxes the need for highly accurate training labels and enables unbiased segmentation. However, the validity of this assumption depends on whether DR can also be extended to the shape domain. This raises the question: does the training set contain enough geometric variability? To account for variations in the morphology across individuals and/or populations, current synthetic methods use templates derived from a large number of subjects automatically processed with FreeSurfer. We identify two major limitations in this strategy.

First, while synthetic training may be robust to random errors at label boundaries, it is not robust to systematic biases. For instance, FreeSurfer predictions of the putamen systematically include part of the claustrum, inducing a consistent bias in the shape of the putamen. Prior work has reported that synthetic models learn these shape priors and consequently reproduce the associated inaccuracies \citep{valabregue_comprehensive_2024}. 

Second, these methods are limited to templates provided by existing automated tools and do not have the flexibility to incorporate additional ones, such as extra-cerebral tissues which are critical in various applications. 

To address these limitations, we propose to leverage domain randomization from a very small subset of subjects with extensive and accurate annotated segmentations. Rather than relying on hundreds of automatically labeled subjects, the approach demonstrates that a limited number of high-quality label templates is sufficient when combined with joint intensity and shape generative modeling. This strategy enables both the addition of new labels and improved control over training label quality. Furthermore, a novel spatial augmentation is introduced to increase shape variability, in particular by modulating cortical thickness at high resolution. In line with the domain randomization principle, this augmentation allows robust generalization to cortical thickness variations despite the small training set.

\section{Contributions}
In this work, we introduce Segment It All Model (SIAM), a whole-head tissue segmentation framework reaching or outperforming SOTA performance on a set of complementary experiments, despite being trained on very few cases. These performances result from the following set of contributions:

\begin{enumerate}
%[leftmargin=0.3 cm]
%\setlength\itemsep{0em}
\item \textbf{More with less.} This work demonstrates that domain randomization can be effectively performed from a very small number of label templates, which makes it possible to learn to segment new anatomical structures. This contrasts with existing approaches that rely on large collections of automatically generated templates, and highlights the importance of annotation quality over quantity.
\item \textbf{Extension of domain randomization to the shape domain.} A novel augmentation strategy is introduced to model anatomical variability. In particular, high-resolution spatial transformations are designed to generate controlled variations in cortical thickness and deep structures, reinforcing model robustness toward tissue volume changes.
\item \textbf{Flexible and extensible labeling framework.} The proposed head segmentation into 16 anatomical structures enables both the correction of systematic biases in existing labels and the integration of additional anatomical structures, including five extra-cerebral classes: CSF, vessels, dura mater, skull, and head. 
\item \textbf{Experimental validation.} Extensive experiments on eight datasets comprising N=301 subjects show that the proposed method achieves performance comparable to existing domain randomization approaches, despite being trained on a very limited number of annotated subjects. Quantitative and qualitative assessments demonstrate less systematic bias relative to other methods, while the modular framework enables addition of new labels and correction of reference-label biases—both of which are critical for accurate volumetric analysis. 
\end{enumerate}
The model, the code for prediction and the label templates used for training are openly available \url{https://github.com/romainVala/SIAM}.

\section{Materials and Methods}
In this section, we present the Segment It All Model (SIAM) framework. We detail the construction of the high-quality, whole-head label templates, the advanced synthetic generative model used to simulate realistic shape and contrast variations, and the network architecture and training procedure.

\subsection{Construction of high-quality training label templates}
\label{label_templates}
As further detailed in the next section, our model is based on the synthetic learning approach in which training data are generated from label templates. To train such a model for segmenting brain and extra-cerebral tissues, an initial dataset with full-head labeling at the tissue level is required. 

We used the MIDA template, the only publicly available template with extra-cerebral labels. In addition, we constructed 5 other templates based on multimodal high-resolution imaging following the same tissues defined in MIDA. Labels for these additional cases were obtained by combining state-of-the-art software for brain segmentation and manual annotation for extra-cerebral structures, across complementary acquisitions.

\subsubsection{The MIDA template (N=1)}
The MIDA template provides whole-head tissue and region segmentation based on manual delineation of a single subject \citep{iacono_mida_2015}. The provided template at 0.5 mm\textsuperscript{3}, includes 116 labels, of which 92 correspond to extra-cerebral tissues.
We regroup the labels into 12 tissues for the brain: gray matter (GM), white matter (WM), cerebellar GM, cerebrospinal fluid (CSF), ventricles, thalamus, putamen, pallidum, caudate, accumbens, amygdala, hippocampus. 11 tissues for extra-cerebral tissues: skin epidermis, head fat, head muscle, salivary glands, air, mucosis, eye ball, skull bone, skull diploë, dura mater, vessel.
\subsubsection{The Skull templates (N=3)}
CT and MRI data were acquired from twelve subjects at the Paris Brain Institute as part of a previous study \cite{bancel_quantitative_2025}. CT scans were performed on a Discovery CT750 HD scanner (GE Healthcare), and MRI scans were acquired on a Siemens Magnetom Prisma 3T scanner. CT images were obtained with a voxel size of $0.49 \times 0.48 \times 0.62$ mm\textsuperscript{3}). Multi-contrast MRI included T1-weighted imaging (MP2RAGE, 1 mm\textsuperscript{3}), UTE (ultra-short echo time, voxel size 0.6 mm\textsuperscript{3}), and FLAIR (1 mm\textsuperscript{3}). All acquisitions were co-registered to the UTE images and resampled to 0.6 mm\textsuperscript{3}). We used three subjects templates for training, and the remaining nine subjects were used for testing skull segmentation precision, with the skull label defined from the CT. Subjects signed a written informed consent approved by the local Ethics Committee (APHP190407 / IDRCB: 2019-A01791-56, ClinicalTrials.gov NCT04074031).

\subsubsection{The Vasculature templates (N=2)}
MRI acquisitions were performed at the Paris Brain Institute on Siemens CIMAX 3T system for two subjects. Multi-sequence acquisitions consisted of T1-weighted MPRAGE (0.7mm\textsuperscript{3}), Dixon sequence (0.7mm\textsuperscript{3}), producing water and fat contrasts, T2-weighted SPACE (0.7 mm\textsuperscript{3}), and phase-contrast MRI (velocity encoding 5 cm/s in three directions, 1.3 mm\textsuperscript{3}). All modalities were co-registered, then resampled to 0.5 mm\textsuperscript{3} resolution. all subjects signed a written informed consent (ID-RCB: 2021-A02404-37).
\subsubsection{Labeling procedure}
Full-head templates were obtained using a semi-automatic approach. First, cerebral labels were initialized using existing segmentation tools, selected based on their respective strengths, with T1-weighted images as input. Specifically, gray, white Matter, amygdala, hippocampus, and ventricle labels were segmented using FreeSurfer. Deep brain nuclei (caudate, accumbens, putamen, pallidum, and thalamus) were obtained using AssemblyNet \citep{coupe_assemblynet_2020}. Cerebellar WM and GM were obtained using DeepCERES (Morell-Ortega et al., 2025). Then, extensive visual assessment and manual editing was performed by a neuroimaging expert with 20 years of experience in brain MRI (R.V.). Only minor corrections were required for these cerebral structures, such as the GM in the occipital region for one subject in the vasculature dataset. 
Additional labels were obtained through manual annotation for each subject: the plexus choroid in the lateral ventricles, the hypophysis, mammillary gland, vessels, dura mater, skull bone, diploë, air cavities, mucosa, eyeballs, muscle, tendon, fat, and epidermis. The cerebrospinal fluid (CSF) label was defined as the space between GM and the skull. This delineation was supported by the multiple contrasts available for each dataset.

We started with the skull dataset, which contains a CT acquisition from which a precise skull delineation is easy to perform by intensity thresholding. We manually segmented the structures mentioned above, except the dura mater and the vessels which were included in the CSF label. To get these specific labels, a first model was trained based on the three subjects to predict the skull on the vasculature dataset. We then completed the labeling of dura mater and vessels on the vasculature dataset and trained another model on the two subjects. This model was then used to predict dura mater and the vessel labels on the skull dataset in order to get the full labeling.

The manual labeling took several weeks per subject, resulting in a limited number of subjects (N=6) that reflects the high cost and expertise required for such detailed anatomical labeling. While such a training set might seem too limited to generalize to various anatomical shapes, we show below how synthetic training can be effective in this context. The next section describes the data generation and training strategy designed to obtain a segmentation model that performs on unseen real data, starting from 6 high-quality label templates.

\begin{figure*}[!t]
  \centering
%  \width
  \includegraphics[width=0.92\textwidth ]{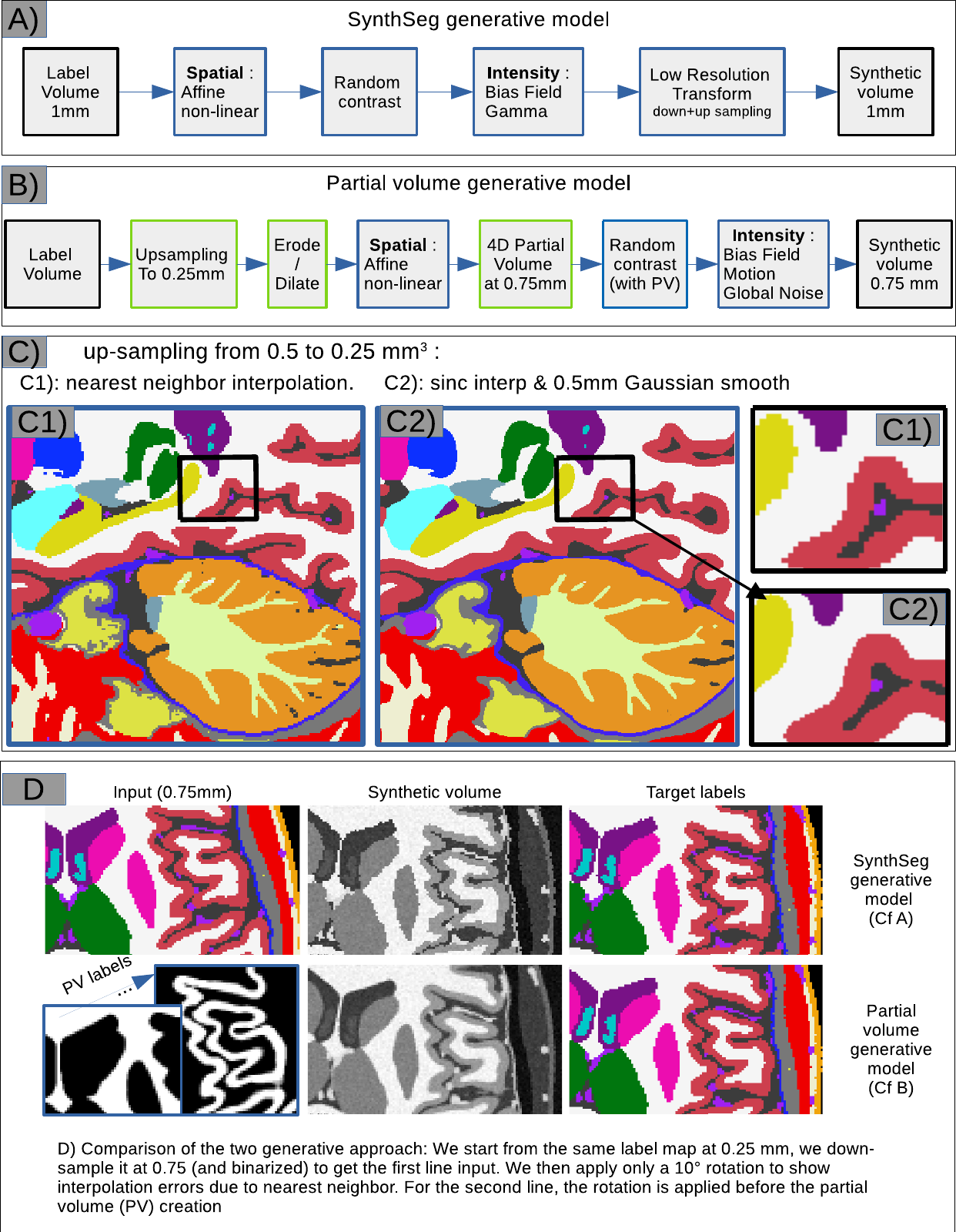}	
  \caption{A) Synthetic data generation as originally proposed by Billot et al. B) Our approach. C) Importance of appropriate high-resolution upsampling: panels C1 and C2 share the same underlying resolution. With nearest-neighbor interpolation (C1), the original 0.5-mm voxel grid remains apparent, with no effective gain in resolution. In contrast, our approach (C2) generates smooth boundaries between structures, revealing an actual resolution enhancement beyond the original voxel grid. D) Benefit of modeling partial volumes for high-quality synthetic data: this enables more realistic transitions in tissue intensity and avoids interpolation artifacts in the target labels, as shown here after a 10 degree in-plane rotation. }
  \label{fig1}
\end{figure*}

\subsection{Synthetic model: from label template to image}
Figure \ref{fig1} A)-B) illustrates the proposed generative model, in comparison with the SynthSeg generative model \citep{billot_synthseg_2023}. The proposed modifications are designed to benefit from the high-resolution label templates. It aims at minimizing interpolation errors, allowing for submillimetric volume changes augmentation. 
\subsubsection{Super resolution of labels template}
 To obtain high-resolution teamplates, all N=6 templates were upsampled to a 0.25 mm isotropic resolution through a three step-procedure. Starting from a label template at any arbitrary resolution, the process is as follows: 1) The label template is converted into a 4D one-hot encoded volume where each channel corresponds to an anatomical structure. Each label channel is then independently upsampled to 0.25 mm isotropic resolution using Sinc interpolation. 2) To reduce interpolation errors from lower-resolution inputs, Gaussian smoothing (0.5 mm kernel) is applied to each channel. 3) The 4D volume is converted back to a 3D binary label template via the argmax function. 
 
Figure \ref{fig1}-C illustrates the benefits of the proposed approach. With nearest-neighbor interpolation (Figure \ref{fig1}-C1), no effective resolution gain is achieved, and the 0.5mm voxel grid remains visible despite the 0.25 mm resolution. In contrast, our approach (Figure \ref{fig1}-C2) produces smoother anatomical boundaries at 0.25 mm. The trade-off is a slight smoothing of fine structures, such as thin cerebellar white matter, cerebrospinal fluid in sulcal fundi, and small vessels.
\subsubsection{Synthetic shape augmentation and resampling}

\paragraph{Labels augmentation erode/dilate:} To increase variability in deep nuclei and GM across the lifespan, random morphological dilations and erosions are applied. Specifically, each tissue is expanded within selected neighboring regions (e.g., white matter dilation within gray matter, ventricles, and deep nuclei; CSF dilation within gray matter and vasculature structures). By varying these combinations, both expansion and shrinkage effects are modeled. The magnitude of the morphological dilation/erosion is randomly varied between 1 and 4 iterations. Working at 0.25 mm resolution is advantageous because it allows generating subtle variations in gray matter thickness, as each iteration corresponds to a 0.25mm extension of the external surface.

\paragraph{Spatial deformation:} Random affine and elastic deformations are applied to the high-resolution (0.25 mm) label templates using TorchIO \cite{perez-garcia_torchio_2021}. Performing these transformations at high resolution reduces interpolation errors compared to applying them at lower resolution (e.g., 0.75 mm), resulting in more accurate training targets, as shown in the third column of Figure \ref{fig1}-D.

\paragraph{4D partial volume at 0.75 mm:} A resolution of 0.75 mm is used for training. Partial volume maps are generated by applying average pooling to the 0.25mm high-resolution 4D one-hot encoded label maps.
\subsubsection{Synthetic contrast augmentation}
\paragraph{Label-to-Image generation:} This step corresponds to the “Random Contrast” component shown in Figure~\ref{fig1}. Following the SynthSeg framework, tissue intensities are sampled from a Gaussian mixture model (means uniformly sampled in [0,1], standard deviations in [0.001, 0.01]). We improve this step by accounting for partial volume effects, following MRI physics: voxel intensities are computed as the weighted sums of tissue-specific signals, with the partial volume maps serving as weights. This allows realistic intensity transitions between tissues, rather than binary boundaries as initially proposed in~\cite{billot_synthseg_2023}. (Figure \ref{fig1}-D).

\paragraph{Intensities transform.} To further enforce the robustness to variations in MRI quality, several intensity-based augmentations are applied, including bias field simulation, random motion artifacts (shown to be critical for neonatal segmentation \citep{valabregue_comprehensive_2024}, and additive Gaussian noise (mean = 0, standard deviation in [0.01, 0.1]). Finally, all images are normalized to the [0,1] intensity range.

% \subsection{Model training and competitors}
\subsection{SIAM network architecture and training}
The training of SIAM was performed using the nnU-Net framework \citep{isensee_nnu-net_2024} with standard settings (1000 epochs, 5-fold ensemble, Dice and cross-entropy losses). The chosen architecture was the 3D full-resolution network with a residual encoder, containing seven blocks (channels: 32, 64, 128, 256, 320, 320, 320). Training used 3D patches of size (256, 256, 192), equivalent to (192, 192, 144) at 1 mm resolution. nnU-Net’s internal data augmentation was disabled since it was handled within our generative model. We precomputed 1000 synthetic image–label pairs offline for each training set and concatenated the three. Predictions used 5-fold ensemble averaging, without post-processing. Training time was approximately two days per fold using an NVIDIA A100-80G GPU. 

\section{Experimental setup}
In this section, we describe the experimental framework used to validate our approach.

\subsection{Test set and reference annotations}
A total of eight datasets, comprising N=301 subjects, were used for the evaluation. Dataset selection was guided by the availability of manual or high-quality reference annotations (REFs). We also included neonates from dHCP and pediatric subjects with anatomical deformations from DBB to assess generalization across a wide age range and a diversity of anatomical shapes.

\textbf{1) MICCAI\_2012 (N=20)} \citep{landman_miccai_2012}: 20 T1-weighted (T1w) scans with manual segmentations from the MICCAI 2012 challenge, acquired at 1 mm isotropic resolution. Segmentations were manually performed by Neuromorphometrics following the BrainCOLOR protocol. Particular care was taken for deep nuclei labeling; however, the use of 2D brush delineation in coronal slices introduces visible step artifacts in other views. Visual inspection also indicates a systematic overestimation of gray matter. Note that an extended version of this dataset is commercially available and has been used for training in prior work \cite{coupe_assemblynet_2020,huo_3d_2019} instead of using FreeSurfer labels.

\textbf{2) Mindboggle (N=101)} \citep{klein_101_2012}: 101 T1w scans acquired at 1 mm isotropic resolution from multiple scanners, publicly available. Although often considered a manually labeled dataset, it is important to note that manual annotation was primarily performed for cortical parcellation. The global GM and deep nuclei labels were automatically computed from FreeSurfer 5.0. Visual inspection reveals systematic errors in deep gray matter structures, consistent with known limitations of FreeSurfer segmentations. As such, we consider this dataset as FreeSurfer-derived silver-standard REF, despite its widespread use as a manual reference.

\textbf{3) DBB (N=37)} \citep{amorosino_dbb-distorted_2022}: 37 T1w scans of pediatric subjects (1–18 years) with congenital or acquired brain abnormalities publicly available. Subjects were divided into two subgroups to isolate the group of four subjects with severe hydrocephalus, referred to as “XXL Ventricles” due to substantially larger anatomical distortions. Segmentations include six tissue classes, with deep gray matter grouped into a single label. All segmentations were performed with the active Contours segmentation mode of ITK-SNAP \cite{yushkevich_user-guided_2019}, and manual correction when necessary. Visual inspection reveals highly variable segmentation quality, with frequent inclusion of dura mater within GM, consistent with limitations of intensity-clustering segmentation methods. 

\textbf{4) Ultracortex (N=12)} \citep{mahler_ultracortex_2025}: 12 T1w scans acquired at 0.6 mm isotropic resolution at 9.4T using MPRAGE or MP2RAGE sequences. The selected subjects (20–53 years) include GM and WM segmentations obtained through extensive manual correction of initial FreeSurfer outputs. Given the high image quality and iterative refinement process, the resulting GM labels are considered of high quality. 

\textbf{5) HCP test-retest (N=82)} \citep{van_essen_wu-minn_2013}: 41 subjects from the Human Connectome Project received two acquisition sessions, using 0.7 mm isotropic T1w and T2w scans acquired at 3T. Data were processed using the HCP minimal preprocessing pipeline, providing high-quality T1/T2 co-registration \citep{glasser_minimal_2013}. GM reference was obtained by running FreeSurfer (v7.4.1, -hires flag) on the preprocessed images, leading to high-quality labels. For test–retest experiments, T1w images from the second session were brain-cropped and co-registered to the first session using NiftyReg \citep{modat_fast_2010}.

\textbf{6) dHCP (N=20)} \citep{edwards_developing_2022}: 20 neonatal subjects from the developing Human Connectome Project, using 0.5 mm isotropic T1w and T2w scans. The 20 oldest healthy subjects (45 weeks post-conception) were selected to ensure sufficient cortical folding. We used the provided GM labels obtained with drawEM \citep{makropoulos_developing_2018}.

\textbf{7) SynthAtrophy (N=20$\mathbf{\times}$7)} \citep{rusak_quantifiable_2022}: 20 subjects with simulated T1w scans generated using a GAN conditioned on gray matter partial volume (PV) maps. The PV maps are derived from FreeSurfer surfaces with controlled levels of cortical thickness atrophy. The dataset is built from 20 healthy subjects from the Alzheimer's Disease Neuroimaging Initiative, with 10 atrophy levels ranging from 0.1 to 1 mm. We selected the subset without atrophy, denoted SynthNoAtrophy, and a subset of six atrophy levels (0.1, 0.3, 0.5, 0.7, 0.9, and 1 mm), leading to 7 scans per subject.

\textbf{8) Skull (private) test set (N=9)}: 9 subjects acquired with the same protocol as the Skull dataset, including co-registered T1-weighted (UNI), FLAIR, UTE, and CT images resampled to 0.6 mm isotropic resolution. This test set is used exclusively for skull evaluation, with labels manually defined from CT scans.

Based on dataset descriptions and careful visual inspection of the provided labels, we revisit the common assumption that all datasets are manually annotated. For deep nuclei, the MICCAI\_2012 dataset provides manual reference annotation, whereas Mindboggle relies on automatically generated labels and should not be considered manual. For GM, only Ultracortex provides consistently high-quality manual delineations. In contrast, other datasets depend heavily on FreeSurfer-derived segmentations and should therefore be regarded as silver-REF. The DBB dataset represents an intermediate case, where manual corrections were applied locally in regions with strong anatomical deformation, particularly where FreeSurfer failed.

\subsection{Competitive models}
We compared our approach to five methods, including the well-established FreeSurfer, its deep-learning variant FastSurfer, and three state-of-the-art synthetic approaches trained on large datasets of FreeSurfer-derived templates.

\textbf{FreeSurfer} \citep{fischl_freesurfer_2012}: FreeSurfer, based on “classical” image processing tools, is the most used segmentation tool, especially for cortical gray matter. Considered as the state of the art, it has been extensively used for training and testing deep learning models.

\textbf{FastSurfer} \citep{henschel_fastsurfervinn_2022}: FastSurfer was trained on more than 1,000 subjects with pairs of real T1w data (1.5T or 3T) and labels automatically generated by FreeSurfer. An interesting advantage of FastSurfer is its interpolation module, which allows processing data at its native resolution. Because FastSurfer was trained on 3T T1-weighted images, it does not generalize well to other contrasts \cite{isensee_automated_2019, valabregue_comprehensive_2024}, and, as shown in this work, underperforms on high-field T1-weighted images.

\textbf{SynthSeg} \citep{billot_synthseg_2023}: SynthSeg was trained using 20 manually segmented templates (from a private dataset containing full-head labels), and 1,000 additional templates from FreeSurfer derived labels. SynthSeg is trained at a resolution of 1 mm.

\textbf{SuperSynth}: SuperSynth is the latest model developed by the SynthSeg team. The model is available through the 8.2.0 version of FreeSurfer. It can be considered an extension of SynthSeg, with predictions still performed at 1 mm resolution, but including additional extra-cerebral structures, similar to those introduced in SAMSEG \citep{puonti_fast_2016}. At the time of this work, it is not associated with a publication, and a detailed technical description is not available.

\textbf{GOUHFI} \citep{fortin_GOUHFI_2025}: GOUFHI is a synthetic approach that operates at a higher resolution (0.7 mm). GOUHFI uses the same generative model as SynthSeg but with templates obtained by FastSurfer on high-resolution acquisitions (N = 206). GOUHFI was trained only on skull-stripped synthetic images, and therefore required a pre-processing step. We use the brain mask tool proposed by the author’s GitHub repository.

\subsection{Experimental setup and metric evaluations.}
%We propose three experiments to evaluate and compare the models:

We propose three evaluation settings to assess segmentation performance and compare the models:
\begin{enumerate}
    \item \textbf{Anatomical accuracy against reference annotations.} Segmentation accuracy was evaluated using the Dice score for each class. Gray matter performance was assessed across the first seven datasets (N = 282 subjects). Subcortical structures were evaluated on MICCAI\_2012 and Mindboggle, while in DBB, deep nuclei were grouped into a single class to match the reference annotations (REF). Skull segmentation was evaluated separately using the dedicated skull test set.

   \item \textbf{Prediction consistency across acquisition protocols.} Prediction consistency across repeated acquisitions of the same subject was assessed by computing Dice overlap between paired segmentations. Two complementary settings were considered: (i) test–retest robustness, comparing two T1-weighted acquisitions acquired in separate sessions; and (ii) cross-contrast robustness, comparing segmentations obtained from paired T1w and T2w scans.

    \item \textbf{Sensitivity to cortical atrophy.} Sensitivity to morphological changes was evaluated on the SynthAtrophy dataset, which provides controlled variations in cortical thickness. In addition to reporting Dice scores and total GM volumes, we quantified the accuracy of atrophy estimation using the relative error in predicted atrophy rates:
\begin{equation} 
\label{eq1}
\mathrm{RelativeAtrophyError} =
1 -
\dfrac{V^{\mathrm{pred}}_{\mathrm{atrophy}}}{V^{\mathrm{pred}}_{\mathrm{baseline}}}
{
\dfrac{V^{\mathrm{ref}}_{\mathrm{baseline}}}{V^{\mathrm{ref}}_{\mathrm{atrophy}}}
}
\end{equation}
where $V^{\mathrm{pred}}_{\mathrm{atrophy}}$ and $V^{\mathrm{pred}}_{\mathrm{baseline}}$ denote the total cortical gray matter volumes estimated by the model for the atrophy and baseline scans, respectively, and $V^{\mathrm{ref}}_{\mathrm{atrophy}}$ and $V^{\mathrm{ref}}_{\mathrm{baseline}}$ are the corresponding volumes derived from the reference annotations. A zero value indicates perfect estimation of the true relative atrophy, positive values indicate model underestimation, and negative values indicate overestimation.
\end{enumerate}

To identify the best-performing models for each task, we conducted a Bonferroni-corrected paired Wilcoxon signed-rank test against the top-scoring model ($p < 0.01$). Bold text indicates the highest mean score and any results that are not significantly different from it.

\section{Results}
In this section, we assess the comparative performance of SIAM and competing models across three settings.

\begin{table*}[!t]
\caption{Gray matter Dice scores ($ \% $) across 7 datasets. Quantitative results similar as in Fig~\ref{fig2}-A.}
\label{tab1}
\centering
\begin{tabular}{lccccccc}
  \toprule
  & \multicolumn{7}{c}{\bf Gray Matter} \\
\cmidrule(lr){2-8}
 & DBB & MICCAI\_2012 & Mindboggle & SynthNoAtrophy & Ultracortex & HCP & dHCP \\
% &  &  &  &  &  &  &  \\
  \midrule
  \rowcolor{LightGray}
  FastSurfer & \textbf{84.0  (5.9)} & 82.5  (2.7) & \textbf{95.1  (0.7)} & 93.6  (0.9) & 73.1  (6.7) & \textbf{96.7  (0.5)} & 26.3  (12.0) \\
SynthSeg & \textbf{84.0  (2.9)} & 86.7  (1.4) & 83.2  (1.4) & 85.2  (2.3) & 80.4  (1.3) & 84.8  (0.9) & 82.4  (0.7) \\
\rowcolor{LightGray}
SuperSynth & \textbf{86.0  (3.2)} & 86.5  (3.1) & 91.3  (0.8) & 92.4  (1.1) & 85.0  (1.0) & 88.5  (0.5) & 86.4  (0.9) \\
GOUHFI & 84.2  (4.2) & \textbf{88.1  (2.1)} & 91.7  (1.3) & \textbf{94.2  (0.7)} & 89.2  (1.5) & 91.7  (0.6) & 89.2  (1.7) \\
\rowcolor{LightGray}
SIAM & \textbf{85.9  (3.8)} & \textbf{87.5  (2.5)} & 91.0  (1.4) & \textbf{94.1  (0.7)} & \textbf{91.6  (1.2)} & 93.8  (0.6) & \textbf{91.2  (1.1)} \\
\bottomrule
\end{tabular}
\end{table*}

\begin{table*}[!t]
\caption{Subcortical Dice scores ($\%$) evaluated on MICCAI\_2012 (manual reference) and Mindboggle (FreeSurfer reference). Quantitative results similar as in Fig~\ref{fig2}-B}.
\label{tab2}
\centering

\resizebox{\textwidth}{!}{
\begin{tabular}{lllllllll}
  \toprule
  & \multicolumn{8}{c}{\bf MICCAI\_2012} \\
\cmidrule(lr){2-9}
  & Putamen & Pallidum & Caudate-Acc & Thalamus & Cerebellum & Ventricle & Hippo & Amygdala \\
  \midrule
  \rowcolor{LightGray}
FastSurfer & 87.9  (1.9) & 78.9  (3.5) & 85.8  (3.8) & 87.6  (1.2) & \textbf{92.2  (1.4)} & 87.2  (5.2) & 81.0  (2.8) & 66.3  (5.5) \\
SynthSeg & 85.1  (1.5) & 84.3  (3.1) & 83.5  (3.1) & 87.7  (1.1) & 91.1  (1.4) & 84.9  (6.1) & 80.6  (2.9) & 65.0  (5.0) \\
\rowcolor{LightGray}
SuperSynth & 86.0  (1.9) & \textbf{86.3  (2.9)} & \textbf{87.5  (2.5)} & 88.3  (1.1) & \textbf{92.5  (1.0)} & \textbf{89.3  (4.3)} & \textbf{81.5  (2.1)} & 65.3  (4.7) \\
GOUHFI & 87.7  (3.1) & 83.4  (4.0) & 79.3  (12.9) & \textbf{89.0  (1.9)} & 88.8  (2.4) & 84.8  (3.2) & \textbf{82.3  (3.1)} & \textbf{68.6  (4.4)} \\
\rowcolor{LightGray}
SIAM & \textbf{90.2  (2.1)} & \textbf{85.3  (2.9)} & \textbf{88.4  (3.2)} & 85.8  (1.3) & 85.3  (1.6) & 87.4  (5.8) & \textbf{83.1  (3.3)} & \textbf{69.7  (5.2)} \\

\midrule
  & \multicolumn{8}{c}{\bf   Mindboggle } \\
\cmidrule(lr){2-9}
  & Putamen & Pallidum & Caudate-Acc & Thalamus & Cerebellum & Ventricle & Hippo & Amygdala \\
% &  &  &  &  &  &  &  &  \\
\midrule
\rowcolor{LightGray}
FastSurfer & 88.0  (1.7) & 78.0  (7.4) & \textbf{89.7  (3.0)} & \textbf{90.7  (1.6)} & \textbf{91.0  (2.5)} & \textbf{92.5  (2.9)} & \textbf{89.3  (2.5)} & \textbf{84.6  (3.5)} \\
SynthSeg & \textbf{89.2  (1.6)} & 81.6  (5.9) & 87.4  (3.3) & 89.4  (1.7) & 87.5  (2.5) & 86.6  (4.2) & 85.1  (2.7) & 80.6  (4.1) \\
\rowcolor{LightGray}
SuperSynth & \textbf{89.0  (1.8)} & \textbf{82.4  (6.1)} & 89.0  (3.2) & 88.0  (1.8) & 89.0  (2.6) & 90.0  (3.3) & 87.9  (2.4) & 83.1  (3.5) \\
GOUHFI & 88.0  (1.9) & 78.7  (7.9) & 88.3  (4.2) & 90.1  (1.6) & 89.1  (2.8) & 89.9  (3.3) & 87.5  (3.2) & 81.0  (4.1) \\
\rowcolor{LightGray}
SIAM & 86.8  (2.3) & 80.5  (5.7) & 87.7  (3.3) & 87.4  (2.0) & 86.2  (2.3) & 87.6  (4.3) & 85.3  (3.2) & 82.0  (3.7) \\

\bottomrule
\end{tabular}
}
\end{table*}

\begin{table}[t]
\caption{Skull Dice scores $(\%)$ when predictions are made from 4 different input contrasts. Quantitative results similar as in Fig~\ref{fig2}-D.}
\label{tab3}
\centering
\resizebox{0.45\textwidth}{!}{\begin{tabular}{lllll}
\toprule
& \multicolumn{4}{c}{\bf Skull } \\
\cmidrule(lr){2-5}
& CT & UTE & FLAIR & UNI \\
\midrule
\rowcolor{LightGray}
SuperSynth & 0.0  (0.0) & 82.9  (2.2) & 82.0  (2.2) & 72.9  (2.8) \\
SIAM & \textbf{91.6  (2.8)} & \textbf{92.2  (1.6)} & \textbf{90.5  (3.0)} & \textbf{88.7  (2.3)} \\

\bottomrule
\end{tabular}}
\end{table}

\begin{figure*}[!t]
  \centering
  \includegraphics[width=\textwidth ]{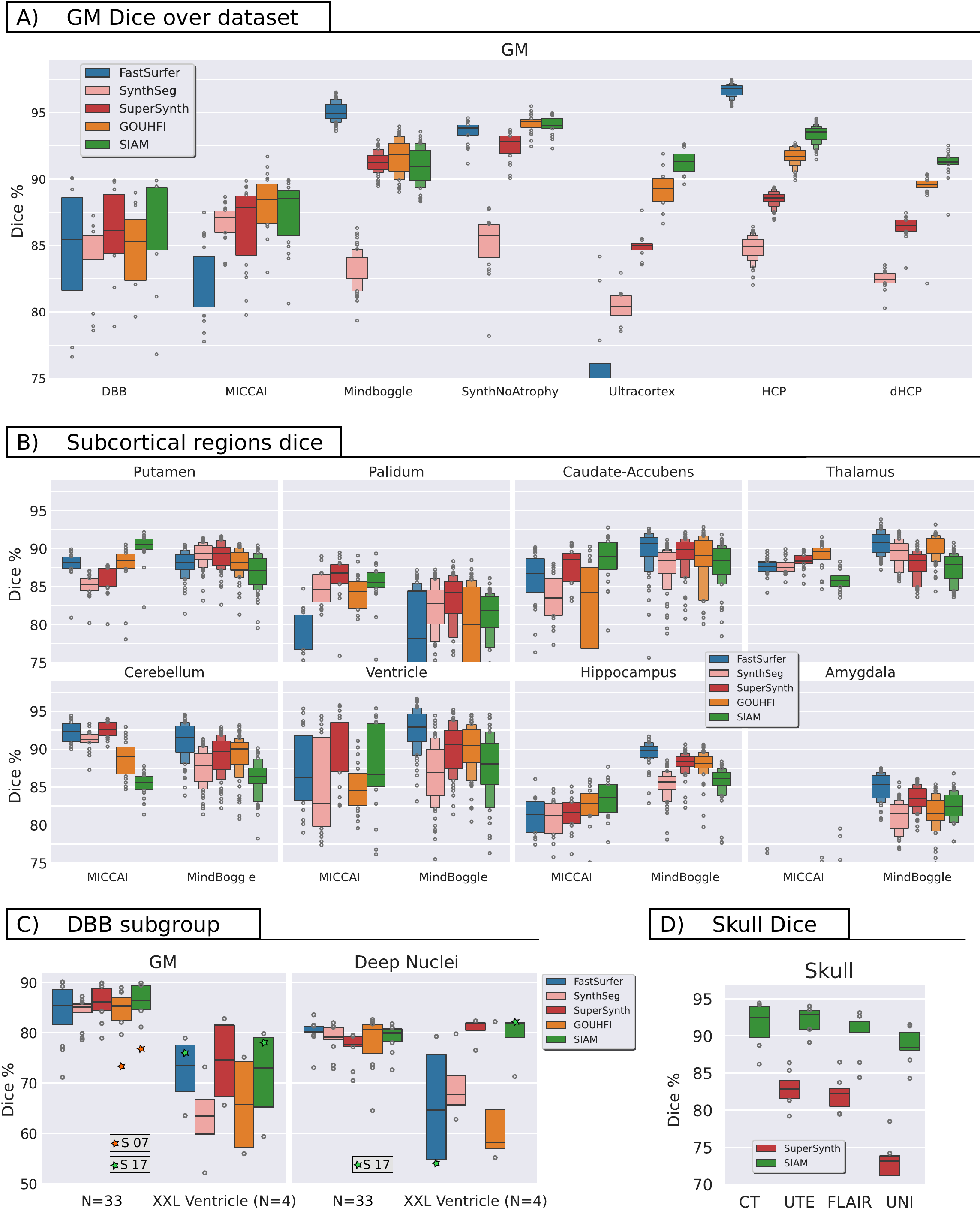}	
  \caption{Anatomical accuracy against reference annotations: (A) Cortical Gray Matter (GM) Dice scores across 7 datasets. (B) Subcortical Dice evaluated on MICCAI\_2012 (manual reference) and Mindboggle (FreeSurfer reference). (C) GM and combined deep nuclei Dice scores on the DBB dataset, separating 4 subjects with severe hydrocephalus (XXL ventricles) from the total cohort. (D) Skull Dice evaluation on the private skull testset when predictions are made from 4 different input contrasts. } 
  \label{fig2}
\end{figure*}
%\clearpage
\begin{figure*}[!t]
  \centering
  \includegraphics[width=0.90\textwidth, keepaspectratio ]{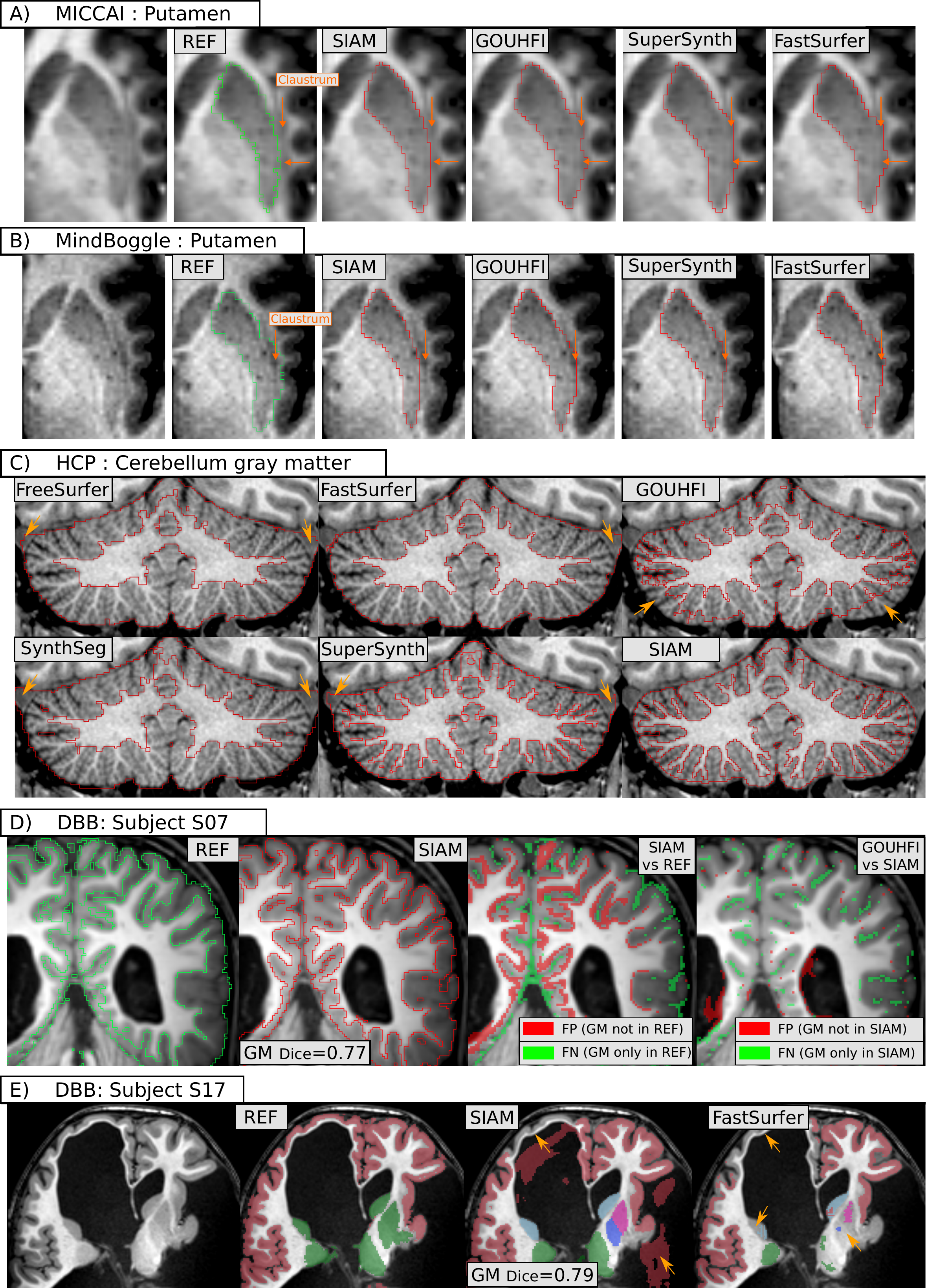}	
  \caption{Qualitative segmentation examples. (A, B) Putamen: all models and the FreeSurfer REF include part of the claustrum, whereas only SIAM and the MICCAI 2012 REF are anatomically correct. (C) Cerebellum: only SIAM captures finer GM/WM details and does not include veins. (D, E) Two DBB outliers with comparable Dice scores but distinct error: sources—reference error in (D) and prediction error in (E).}
  \label{fig3}
\end{figure*}

\begin{table*}[!t]
\caption{Prediction consistency: Dice scores ($\%$) evaluation in Test/Retest. Quantitative results similar as in Fig~\ref{fig4}-A}
\label{tab4}
\centering

\resizebox{\textwidth}{!}{
  \begin{tabular}{lccccccccc}
\toprule  
& \multicolumn{9}{c}{\bf T1 vs T1 repeated } \\
\cmidrule(rl){2-10}
& GM & Putamen & Pallidum & Caudate-Acc & Thalamus & Cerebellum & Ventricle & Hippo & Amygdala \\
\midrule
\rowcolor{LightGray}
FreeSurfer & 92.9  (1.6) & 92.6  (2.8) & 87.3  (9.0) & 86.1  (3.8) & 94.8  (1.2) & 93.5  (2.7) & 93.2  (1.4) & 91.9  (1.9) & 90.5  (1.5) \\
FastSurfer & 93.4  (1.7) & 96.3  (0.7) & 95.0  (1.2) & 94.8  (1.0) & 97.3  (0.4) & \textbf{96.5  (0.9)} & 95.2  (1.1) & 95.7  (0.7) & 92.6  (2.1) \\
\rowcolor{LightGray}
SynthSeg & 93.6  (0.8) & 95.3  (0.4) & 93.4  (0.7) & 91.6  (1.2) & 96.6  (0.2) & 96.3  (0.5) & 92.4  (1.4) & 93.7  (0.5) & 93.7  (0.7) \\
SuperSynth & \textbf{95.5  (1.5)} & \textbf{97.8  (0.6)} & \textbf{97.4  (0.8)} & \textbf{97.0  (1.0)} & \textbf{98.5  (0.3)} & \textbf{96.8  (0.9)} & \textbf{96.5  (0.9)} & \textbf{97.1  (0.6)} & \textbf{95.9  (1.6)} \\
\rowcolor{LightGray}
GOUHFI & 91.8  (1.5) & 95.5  (0.6) & 92.6  (1.5) & 93.5  (0.8) & 95.4  (0.9) & 94.3  (0.8) & 93.7  (1.2) & 94.6  (0.8) & 92.3  (1.4) \\
SIAM & \textbf{95.4  (1.4)} & 97.6  (0.6) & 96.5  (0.8) & 95.6  (1.7) & 97.6  (0.4) & 95.3  (1.3) & 95.8  (0.9) & 96.8  (0.6) & \textbf{96.4  (0.7)} \\

\midrule
& \multicolumn{9}{c}{\bf T1 vs T2} \\
\cmidrule(rl){2-10}
& GM & Putamen & Pallidum & Caudate-Acc & Thalamus & Cerebellum & Ventricle & Hippo & Amygdala \\
%&  &  &  &  &  &  &  &  &  \\
\midrule
\rowcolor{LightGray}
FreeSurfer & -- & -- & -- & -- & -- & -- & -- & -- & -- \\
FastSurfer & -- & -- & -- & -- & -- & -- & -- & -- & -- \\
\rowcolor{LightGray}
SynthSeg & 91.7  (0.4) & 95.4  (0.9) & 92.9  (0.9) & 91.6  (1.5) & 94.3  (0.8) & 93.0  (0.8) & 92.8  (1.7) & 93.3  (0.7) & 92.4  (1.1) \\
SuperSynth & 92.7  (0.4) & \textbf{96.1  (0.7)} & \textbf{94.4  (0.9)} & 93.5  (1.3) & \textbf{96.3  (0.5)} & \textbf{94.5  (0.6)} & 93.8  (1.4) & 94.0  (0.6) & 92.7  (1.1) \\
\rowcolor{LightGray}
GOUHFI & 90.2  (1.0) & 94.5  (1.0) & 89.4  (1.2) & 90.8  (1.2) & 94.3  (0.7) & 90.7  (1.0) & 90.4  (2.8) & 92.5  (0.7) & 92.4  (1.0) \\
SIAM & \textbf{95.2  (0.4)} & \textbf{96.1  (0.9)} & 92.7  (1.2) & \textbf{94.3  (1.3)} & 94.5  (0.7) & 93.7  (0.6) & \textbf{94.5  (1.3)} & \textbf{95.1  (0.6)} & \textbf{94.9  (0.8)} \\

\bottomrule
\end{tabular}
}
\end{table*}

\begin{table*}[!t]
\caption{Sensitivity to gray matter atrophy: Relative atrophy rate prediction errors ($\%$). Same as Fig~\ref{fig4}-D}
\label{tab5}
\centering
\begin{tabular}{lccccccc}
\toprule
& \multicolumn{7}{c}{\bf Atrophy level } \\
\cmidrule(rl){2-8}
& All & 0.1 & 0.3 & 0.5 & 0.7 & 0.9 & 1 \\
\midrule
\rowcolor{LightGray}
FreeSurfer & \textbf{23.6  (11.0)}  & \textbf{31.2  (19.7)} & \textbf{19.9  (9.3)} & \textbf{22.9  (7.3)} & \textbf{23.6  (7.8)} & \textbf{21.8  (6.5)} & \textbf{22.2  (7.2)} \\
FastSurfer &\textbf{24.5  (5.5)} & \textbf{27.1  (9.7)} & \textbf{24.8  (5.7)} & \textbf{24.2  (3.9)} & \textbf{23.8  (2.3)} & \textbf{23.0  (2.6)} & \textbf{24.0  (5.1)} \\
\rowcolor{LightGray}
SynthSeg & 68.3  (7.7)   & 64.7  (10.3) & 65.0  (5.9) & 66.4  (5.3) & 68.8  (5.4) & 71.1  (6.2) & 73.7  (8.4) \\
SuperSynth & 45.7  (6.5)  & 48.0  (11.3) & 46.5  (6.4) & 45.9  (5.2) & 45.3  (4.4) & 44.1  (3.9) & 44.1  (4.9) \\
\rowcolor{LightGray}
GOUHFI &  30.6  (10.1) & \textbf{28.4  (21.0)} & \textbf{28.6  (8.0)} & \textbf{29.1  (5.4)} & 30.8  (3.9) & 32.4  (4.6) & 34.5  (5.9) \\
SIAM &\textbf{24.0  (5.0)} & \textbf{27.2  (7.8)} & \textbf{26.0  (4.4)} & \textbf{23.7  (3.9)} & \textbf{22.7  (3.1)} & \textbf{21.9  (3.6)} & \textbf{22.3  (4.2)} \\
\bottomrule
\end{tabular}
\end{table*}

\begin{figure*}[!t]
  \centering
  \includegraphics[width=0.89\textwidth ]{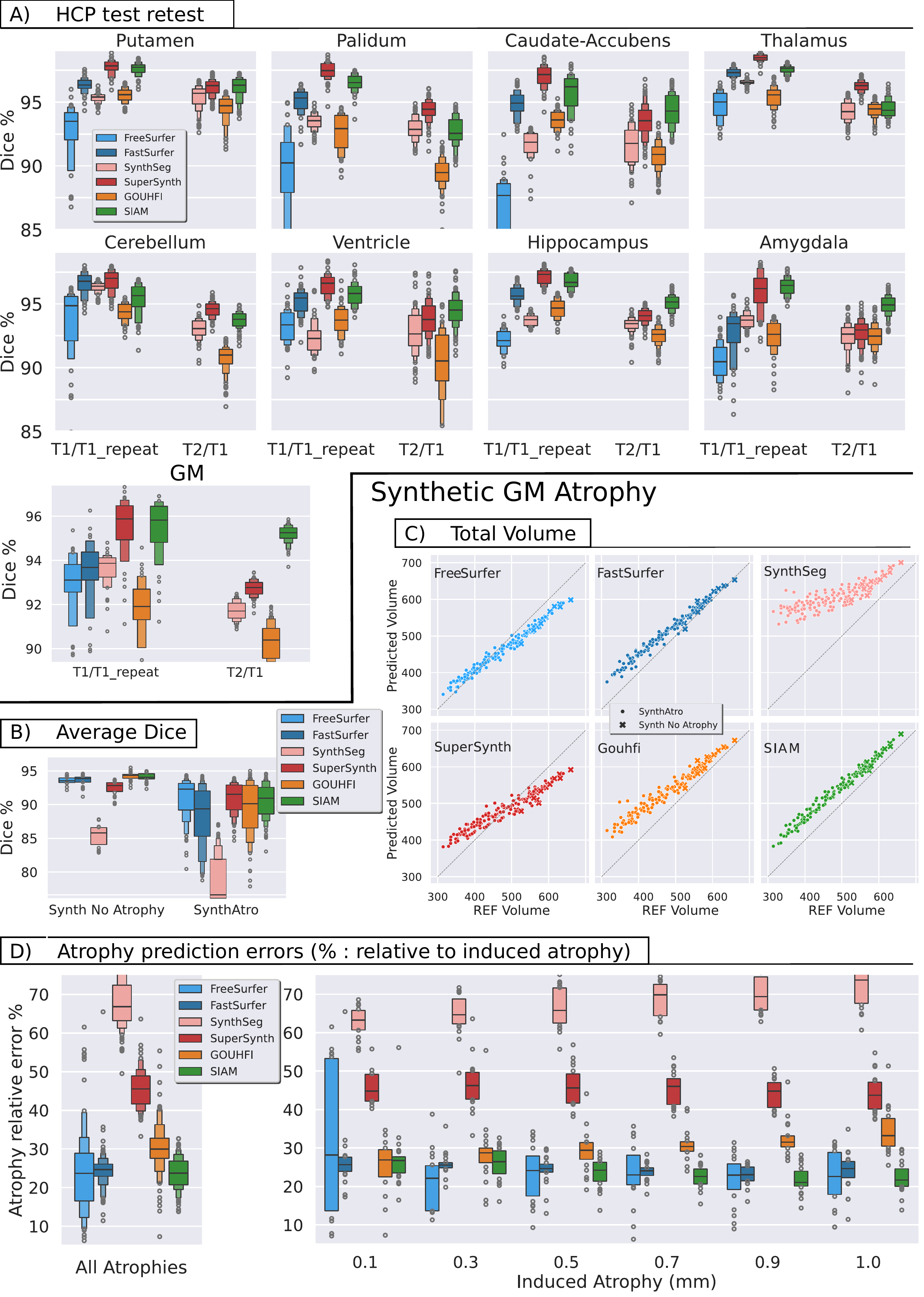}	
  \caption{(A) Prediction consistency: Dice evaluation on the HCP test set, comparing T1w versus T1w\_repeat and T1w versus T2w. (B–D) Sensitivity to GM atrophy: (B) average Dice scores for subjects with and without atrophy; (C) absolute predicted volumes (with reference volume marked as $\times$; (D) relative atrophy prediction errors as defined in \ref{eq1}. Despite similar average Dice scores, the accuracy of relative atrophy prediction differs markedly across models.} 
  \label{fig4}
\end{figure*}

\begin{figure*}[!t]
  \centering
  \includegraphics[width=\textwidth ]{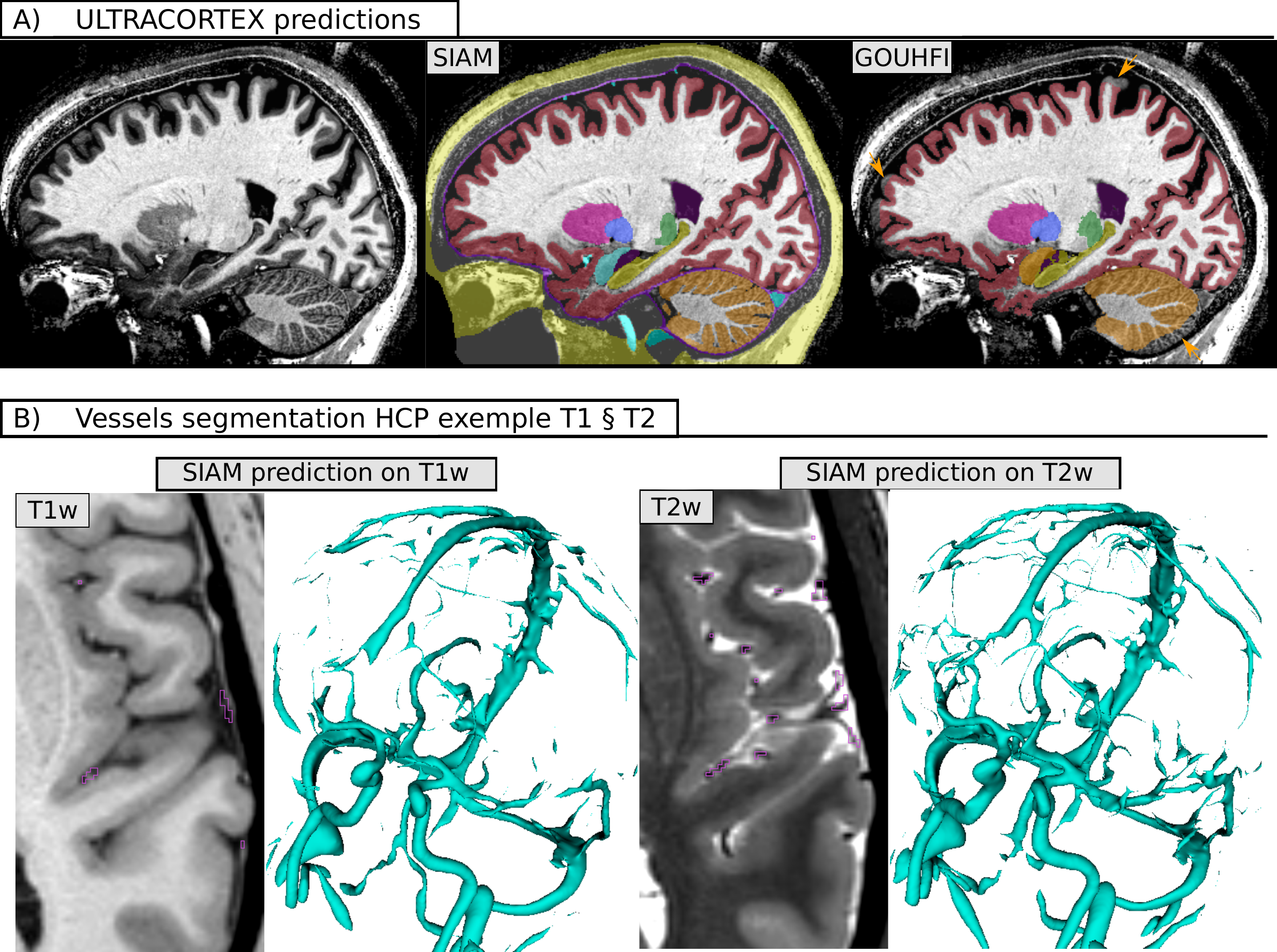}	
  \caption{Qualitative segmentation examples. (A) ULTRACORTEX examples, where the GOUHFI brain mask erodes part of the GM (yellow arrows). (B) HCP examples of vessel segmentation. T2w predictions are denser, as small vessels are more visible on this sequence.} 
  \label{fig5}
\end{figure*}

\subsection{Anatomical accuracy against reference annotations} 

We evaluate Dice scores relative to the corresponding reference annotations on eight test sets covering a large age range from newborns to adults, including subjects with brain-shape distortions induced by pathology.

Figure \ref{fig2}A and Table \ref{tab1} show Dice scores for GM across four datasets with manual reference annotations, one simulated dataset and two datasets with silver-standard REF. We observe large variations in the performance of all models across datasets, with differing rankings across competitive models:
1) FastSurfer outperforms all synthetic models when evaluated on HCP (97\%) and Mindboggle (95\%), for which the reference was obtained using FreeSurfer. On DBB, MICCAI\_2012 and SynthAtrophy, it performs similarly to others, but it fails on the Ultracortex and dHCP datasets.
2) Among the synthetic models, with the exception of SynthSeg which generally underperforms, the methods achieve comparable Dice scores on the 1-mm resolution datasets (MICCAI\_2012, DBB, Mindboggle, and SynthNoAtrophy). In contrast, on the high-resolution datasets (Ultracortex, HCP, and dHCP), SIAM significantly outperforms the other synthetic approaches. We observe a clear sequential progression: SynthSeg reaches the lowest performance (80.4\%, 84.8\%, 82.4\%), followed by SuperSynth (85.0\%, 88.5\%, 86.4\%), GOUHFI (89.2\%, 91.7\%, 89.2\%) and SIAM (91.6\%, 93.8\%, and 91.2\%)

% For synthetic models, except for SynthSeg which underperforms, models show similar Dice scores on the 1-mm-resolution datasets (MICCAI\_2012, DBB, Mindboggle and SynthAtrophy). For high-resolution datasets (Ultracortex, HCP and dHCP), we observe a clear ordering: SynthSeg (80 / 85 / 82) $<$ SuperSynth (85 / 89 / 86) $<$ GOUHFI (89 / 92 / 89) $<$ SIAM (91 / 93 / 91).

Figure \ref{fig2}B and Table \ref{tab2} show Dice scores for subcortical regions on the two datasets where a REF is available (MICCAI\_2012 and Mindboggle). Although performance variations across models are smaller, we also observe a different ranking of the models depending on the dataset. This is particularly clear for the putamen, where SIAM is the best-performing model on MICCAI\_2012 but the worst on Mindboggle. Recall that the first one is a manual reference whereas the second one is obtained with FreeSurfer.

Figure \ref{fig2}C extends the evaluation on the DBB dataset by separating the four subjects with extremely large ventricles, as they represent extreme anatomical deformations. For GM, all models partially fail on this subgroup. For the deep nuclei, only SIAM and SuperSynth show performance similar to that reached on the other 33 subjects, demonstrating their robustness to large anatomical deformations.

Figure \ref{fig2}D and Table \ref{tab3} show the evaluation on the skull dataset, on 9 subjects with 4 different contrasts. SIAM achieves the highest Dice scores, with better results on CT (91.6\%) and UTE (92.2\%) contrasts. Surprisingly, SuperSynth fails to segment CT images and achieves lower performance on the other MRI contrasts (82.9\% on UTE).

The validity of such quantitative analysis assume that the reference labels are anatomically accurate. Visual inspection is therefore essential, and we illustrate representative observations in Figure~\ref{fig3}. We observe differing reference definitions for the putamen: for the Mindboggle dataset, the REF shows a typical FreeSurfer delineation of the putamen, in which part of the claustrum is systematically included (Figure \ref{fig3}-A). In contrast, the manual delineation available for the MICCAI\_2012 dataset does not include the claustrum (Figure \ref{fig3}-B). Regarding model predictions, only SIAM achieves superior putamen segmentation, without partially including the claustrum. These systematic differences, visible across all subjects, explain the ranking of the models for putamen Dice scores: SIAM achieves the highest Dice score when compared with the REF from MICCAI\_2012, and the lowest when compared with the REF from Mindboggle.

Figure~\ref{fig3}-C displays segmentation examples on cerebellar GM. SIAM more closely follows fine anatomical details, leading to a cleaner segmentation, yet one that differs from the available REF, explaining its lower Dice scores across all datasets. Figure~\ref{fig3}-D shows an example of a low Dice score due to large errors in the silver-standard reference from FreeSurfer. Figure~\ref{fig3}-E shows another outlier with a similarly low Dice score, but in this case it is due to prediction errors in regions where cortical geometry is strongly affected by massive expansion of the ventricles. Figure~\ref{fig3}-E also demonstrates superior segmentation of deep nuclei by SIAM in this challenging case.

\subsection{Prediction consistency across acquisition protocols}
Figure~\ref{fig4} and Table~\ref{tab4} present results for the HCP test–retest dataset (41 subjects with 2 sessions). We compute Dice scores between predictions obtained from the two acquisitions of each subject (N=41). We also evaluate Dice scores between predictions obtained from T1w and those obtained from T2w images acquired during the same session (N=82). FastSurfer and FreeSurfer are excluded from the latter comparison, since they cannot predict segmentations from T2w volumes.

The general trend is that SuperSynth and SIAM are more consistent than the others, whereas FreeSurfer performs worst for all subcortical regions. FastSurfer improves upon FreeSurfer on all subcortical regions but does not surpass synthetic models. Among synthetic models, GOUHFI reaches the worst performance. Finally, SIAM shows the highest T1w/T2w similarity for GM, with a Dice score close to that obtained in the T1w test–retest evaluation, whereas SynthSeg and SuperSynth experience a drop in performance.

\subsection{Sensitivity to cortical atrophy}
To assess sensitivity to cortical changes, we utilized the SynthAtrophy dataset, consisting of T1w scans from 20 subjects with six simulated levels of GM atrophy, generated by decreasing cortical thickness from 0.1 to 1 mm.

Figure~\ref{fig4}-B shows the average Dice score across subjects at all atrophy levels. All models, except SynthSeg, achieve equivalent scores, with lower performance observed in the presence of atrophy.

Figure~\ref{fig4}-C shows total GM volume predictions as a function of REF values. Different trends emerge for the models’ absolute volume predictions. All models exhibit greater overestimation for smaller GM volumes, and only SuperSynth and FreeSurfer show underestimation in subjects without atrophy. Finally, Figure~\ref{fig4}-D and Table~\ref{tab5} show relative atrophy-rate errors. SynthSeg and GOUHFI exhibit increasing errors with higher atrophy levels, whereas SIAM and FastSurfer achieve the best performance with a relative error of $\simeq 25\%$.

Figure~\ref{fig5} shows the anatomical accuracy of SIAM segmentations for the three extracerebral tissues added to our task (vessels, dura mater, and skull). Notably, full-head labeling enables SIAM to process data without any preprocessing, whereas GOUHFI must rely on an external tool to produce a brain mask. In the representative example shown in Figure~\ref{fig5}-A, the brain mask computed by GOUHFI was too restrictive, leading to missed GM. In contrast, SIAM precisely segments the dura mater, skull, and other head tissues. Figure~\ref{fig5}-B shows vessel predictions obtained from one subject from the HCP test set. In the T2w contrast, small veins within the CSF are clearly visible, but not in the T1w. Predictions from the T2w volume contain more vessels than those from the T1w, yet small vessels that are not visible on this sequence remain undetected.

\section{Discussion}
Despite employing a limited number of templates (N=6), SIAM achieves performance comparable to, and in several cases exceeding, state-of-the-art synthetic models trained on large collections of automatically labeled data. These results support an alternative paradigm in which segmentation quality is driven less by dataset size than by the fidelity of anatomical priors and the control of variability through synthetic generation. By combining high-quality annotations with joint intensity and shape domain randomization, SIAM not only ensures generalization across contrasts and resolutions, but also improves sensitivity to subtle anatomical variations such as cortical thickness.

The discussion is organized around four main aspects: (i) the impact of reference annotation quality on the interpretation of accuracy, (ii) the evaluation of consistency and sensitivity to volume changes beyond standard metrics, (iii) the role and limitations of the learned spatial priors, and (iv) the implications of whole-head modeling for reducing the reality gap and extending segmentation to more comprehensive anatomical representations.
\subsection{Accuracy : biased predictions or biased reference annotations ?}
Quantitative evaluation of segmentation accuracy fundamentally relies on the assumption that reference annotations (REF) provide a valid ground truth. However, as highlighted in prior work \citep{dorent_learning_2021,jannin_model_2006,maier-hein_metrics_2024,sisic_deep_2025}, this assumption may not hold in practice, leading to potentially misleading comparisons between methods. Our results highlight how the definition and quality of the reference annotations (REF) directly impacts evaluation and can bias comparisons.

The large variations in Dice scores across datasets for gray matter likely stem from differences in the quality of reference labels rather than actual differences in model performance. Although manual annotations are often considered the gold-standard REF, they do not necessarily guarantee higher accuracy. In practice, manual delineation of large and complex structures such as GM is typically initialized from automated segmentations, whose systematic biases are difficult to fully correct.

Conversely, although not manually annotated and known to exhibit systematic biases, FreeSurfer-derived GM segmentations can achieve high quality on high-resolution datasets. This is especially the case for submillimeter data such as UltraCortex, HCP, and dHCP. Interestingly, SIAM outperforms all other methods on these high-resolution datasets, which suggests that it may better capture fine anatomical details.

Beyond quality in high-resolution, FreeSurfer generalizes relatively well across acquisition protocols of T1w imaging. FreeSurfer-derived annotations are therefore a practical reference for evaluating generalization capabilities. FastSurfer shows a clear performance gap between datasets similar to its training data (Mindboggle and HCP) and those with different contrasts (UltraCortex and dHCP), highlighting limited generalization. In contrast, synthetic approaches obtain consistent performance across datasets, highlighting their generalization capabilities.

More generally, models that reproduce the same biases or labeling conventions as the reference annotations may achieve artificially high Dice scores. For example, the higher performance of FastSurfer for GM on Mindboggle and HCP is likely due to an overfitting of the model toward FreeSurfer systematic error. A similar effect is observed for subcortical structures. For the putamen, synthetic models trained on FreeSurfer-derived templates outperform SIAM on Mindboggle (FreeSurfer-based references), whereas SIAM outperforms them on MICCAI\_2012 (manual references). This discrepancy is consistent with known biases in FreeSurfer segmentations for deep nuclei \citep{manjon_volbrain_2016,patenaude_bayesian_2011-1}, and illustrates how differences in annotation protocols can drive apparent performance variations.

In other cases, interpretation is more challenging, as manual and FreeSurfer-derived REFs may share similar biases. In the cerebellum, we use a label derived from DeepCeres, which provides finer anatomical delineation. As illustrated in Figure \ref{fig3}-C, SIAM obtains lower Dice scores despite improved segmentation quality, particularly at the white matter boundary. This suggests that, when reference annotations are themselves imperfect, quantitative metrics alone may not reliably reflect anatomical accuracy.

Overall, these findings emphasize that segmentation accuracy cannot be interpreted independently of the reference annotations. Careful visual inspection of both predictions and labels is therefore essential to properly assess model performance. In particular, lower agreement with imperfect references may reflect more anatomically accurate segmentations, whereas higher scores may indicate agreement with biased annotations. SIAM illustrates this behavior, achieving lower Dice scores on lower-quality references (e.g., cerebellum) but stronger performance on higher-quality annotations (e.g., GM in high-resolution datasets and putamen in MICCAI\_2012).
\subsection{Beyond Accuracy: Consistency and Sensitivity to Volume Changes}
While accuracy relative to a reference annotation is the most commonly reported metric, it does not fully capture the properties required for most neuroimaging applications. In many settings, such as longitudinal studies or group comparisons, the ability of a model to produce consistent measurements and to reliably capture relative volume changes can be of greater importance than absolute agreement with a potentially biased reference. 

The HCP test-retest results confirm that most models achieve good consistency, with mean Dice scores above 95\% across methods. In particular, SIAM and SuperSynth achieve the highest average mean Dice score with an average of 96.3\% and 96.9\% over 9 evaluated structures. In contrast, variability remains higher for FreeSurfer, particularly in deep nuclei, consistent with its known limitations \citep{manjon_volbrain_2016,patenaude_bayesian_2011-1}. An interesting observation is that GHOUFI, which outperformed SynthSeg in terms of accuracy on HCP, shows lower consistency for GM. This suggests that SynthSeg may produce systematically biased segmentations that, while less accurate, are more consistent across repeated acquisitions.

For the robustness against contrast changes (T1 versus T2), FreeSurfer and FastSurfer are discarded because they can not segment T2w images. Depending on the region, SIAM or SuperSynth are the best performing approaches, and it is worth noting that for GM, SIAM (95.2\% Dice) better performs than SuperSynth (92.7\% Dice).

An important observation is that consistency and accuracy are not necessarily aligned. For instance, models that reproduce systematic biases in the reference annotations may achieve high Dice scores while maintaining consistent but biased predictions. This is illustrated by the similarity between FreeSurfer and FastSurfer. While FastSurfer achieves the highest accuracy when evaluated against FreeSurfer-derived GM references (mean Dice: 97\%), both methods obtain relatively lower consistency in the test-retest setting (mean Dice: 93\%). This indicates that their predictions and their errors are highly correlated.

To further assess sensitivity to volume changes, we use a synthetic dataset with known atrophy levels, allowing precise quantification of prediction errors. As shown in Fig. D, all methods obtain substantial atrophy errors, highlighting the difficulty of the task. SIAM achieves the lowest and most stable errors across atrophy levels, performing comparably to or better than FastSurfer. In contrast, other synthetic models show larger errors, with SynthSeg exhibiting the highest deviations.

These findings complement previous studies that evaluated sensitivity indirectly through population differences \citep{billot_synthseg_2023,fortin_GOUHFI_2025}. For example, SynthSeg reported effect sizes comparable to FreeSurfer for detecting hippocampal atrophy in Alzheimer’s disease \citep{billot_synthseg_2023}. In this study, we instead found that SynthSeg is less sensitive to changes. This could be explained by differences in evaluation settings (e.g., global GM versus localized structures such as the hippocampus) and by the controlled nature of the synthetic experiment, which amplifies differences between methods. 
Overall, these results highlight that both consistency and sensitivity to volume changes must be explicitly evaluated.

\subsection{Spatial priors and limitations}
Brain segmentation inherently relies on spatial priors. While classical Bayesian methods use an explicit prior from population templates \citep{ashburner_unified_2005,avants_open_2011,puonti_accurate_2020}, deep learning methods implicitly learn them from training data. In both cases, the model's ability to generalize to unseen anatomies critically depends on how well these priors capture the variability of brain structures. 

A common strategy to promote generalization is to rely on large training datasets, with the assumption that anatomical variability is sufficiently represented. In contrast, our approach adopts a different perspective by leveraging a limited number of high-quality templates and explicitly shaping the variability through label-based augmentations. This strategy requires careful design of the augmentation process, but it provides direct control over the anatomical distribution seen during training. In particular, the proposed high-resolution erosion–dilation scheme enables subtle and continuous variations in cortical thickness.

While strong performance is observed for gray matter, other structures such as ventricles, vessels, and dura mater may still be constrained by a limited spatial prior. Further work will focus on better capturing their shape variability. A broader limitation, shared with most current methods, is limited generalization to pathologies inducing strong changes in the geometry of anatomical regions. Nevertheless, synthetic training remains a promising framework to jointly learn healthy tissue and lesions, as explored in prior work \citep{billot_joint_2021,chalcroft_synthetic_2025,lhermitte_synthetic_2025}.
\subsection{Segment it all !}
An important obstacle for the adoption of synthetic training strategy lies in the so-called “reality gap” \citep{jakobi_noise_1995}, referring to the discrepancy between synthetically generated images and real MRI data. While domain randomization has proven effective for achieving contrast-agnostic models \citep{billot_synthseg_2023,fortin_GOUHFI_2025,valabregue_comprehensive_2024}, it does not address the absence of anatomical structures that are not included in the label templates. This omission constitutes a source of mismatch, as unmodeled tissues are simply not represented in the generative process. In this work, we mitigate this limitation by explicitly adding vessels and extra-cerebral tissues, thereby improving the anatomical completeness of the synthetic images.

This more comprehensive modeling has direct consequences on segmentation quality. In standard pipelines, tissues not explicitly represented in the label space are often segmented into neighboring structures with similar intensities. For instance, the dura mater is frequently segmented as gray matter in T1-weighted images. We observe similar behavior in FreeSurfer, FastSurfer, and GOUHFI, whereas SIAM is less prone to such errors due to the explicit representation of surrounding tissues. By reducing these ambiguities, whole-head modeling leads to more anatomically consistent segmentations.

Another important advantage of this approach is the removal of preprocessing steps such as brain extraction. Methods relying on skull-stripped inputs depend on the quality of external brain masks, which can introduce errors and reduce robustness, particularly in challenging datasets. As an example, GOUHFI relies on FastSurfer-derived labels and requires a brain extraction step prior to segmentation. As shown on the Ultracortex dataset, this additional pre-processing step introduces errors due to overly restrictive brain masks. By modeling main head tissues, SIAM achieves improved GM predictions in such cases. Moreover, accurate skull segmentation enables better estimation of intracranial volume, an important normalization factor for downstream volumetric analyses.

Beyond technical improvements, extending segmentation to whole-head anatomy opens new perspectives for clinical and research applications. While most neuroimaging studies focus on brain tissues, extra-cerebral structures also carry relevant information. For example, temporalis muscle thickness may serve as a surrogate to assess sarcopenia in patients with glioblastoma \citep{sadhwani_temporal_2022}, the skull is of importance for transcranial brain stimulation \citep{diedrichsen_modeling_2025}, or for focused ultrasound \citep{manuel_ultra-short_2025}. More generally, the ability to jointly segment brain and non-brain tissues enables a more integrated representation of head anatomy.

Finally, this framework provides a flexible basis for extending current labeling schemes. Even within the brain, FreeSurfer labeling schemes remain incomplete. For example, brainstem nuclei that are only visible on T2w image are missing. Only a limited number of methods currently address structures such as the red nucleus or substantia nigra \citep{bazin_multi-contrast_2020,casamitjana_probabilistic_2025,saranathan_comprehensive_2025}. By relying on a small number of high-quality templates, additional structures can be incrementally incorporated into the model, without the need for large-scale re-annotation efforts. In this sense, synthetic training not only improves segmentation performance, but also offers a scalable pathway toward more comprehensive and anatomically faithful representations.

\section{Conclusion}
In this work, we introduced the Segment It All Model (SIAM), a contrast-agnostic 3D whole-head segmentation framework trained entirely on synthetic data derived from only six high-quality manual templates. By extending domain randomization to the shape domain through high-resolution spatial and morphological augmentations, SIAM successfully overcomes the systematic biases inherent in models trained on large collections of automated ``silver-standard'' labels. 
Extensive evaluations across diverse datasets demonstrate that SIAM achieves state-of-the-art segmentation performance, improved consistency across multi-contrast and test-retest acquisitions, and high sensitivity to cortical atrophy. Furthermore, by explicitly modeling extra-cerebral tissues, such as the skull, dura mater, and vessels, SIAM eliminates the need for error-prone preprocessing steps like skull-stripping. Ultimately, our synthetic training approach allows to prioritize annotation quality over quantity, leading to an unbiased and easily extensible framework for brain image segmentation.

\section*{Acknowledgment}
We would like to thank Dr Nadya Pyatigorskaya and Dr Elodie Hainque for providing the skull dataset which was funded by Recherche" under the program ‘‘Future Investments’’ with the reference ANR-10-EQPX-15, IAIHU-06 (Paris Institute of Neurosciences – IHU), and ANR-11-INBS-0006. This work was performed using HPC resources from GENCI–IDRIS (Grant 2022-AD011011735R3). 
The research leading to these results has received funding from Agence Nationale de la Recherche as part of the ``France 2030'' program (reference ANR-23-IACL-0008, PRAIRIE-PSAI) and as part of the ``Investissements d’avenir'' program (reference ANR-19-P3IA-0001, PRAIRIE 3IA Institute; and reference ANR-10-IAIHU-0006). The ARAMIS Lab is affiliated with DIM C-BRAINS, funded by the Conseil Régional d’Ile-de-France. R.D. received a Marie Sklodowska-Curie grant No 101154248 (project: SafeREG).
Developing Human Connectome Project, Grant/Award Number: ERC 319456; European Research Council, Grant/Award Number: 319456
Thanks to  Human Connectome Project, WU-Minn Consortium (Principal Investigators: David Van Essen and Kamil Ugurbil; 1U54MH091657) funded by the 16 NIH Institutes and Centers that support the NIH Blueprint for Neuroscience Research;

\bibliographystyle{apacite}
\bibliography{lib_zotero}

\onecolumn

% Force Appendix S = 19, reset and reformat figure numbers.
%\setcounter{section}{19}
%\setcounter{figure}{0}
%\renewcommand{\thefigure}{\thesection\arabic{figure}}

%\section*{Supplementary material}
%\label{sec:supplement}

\end{document}